\def\paperID{4163}
\def\confName{CVPR}
\def\confYear{2025}

\def\paperTitle{GCC: Generative Color Constancy via Diffusing a Color Checker}

\def\authorBlock{
Chen-Wei Chang$^1$
\quad
Cheng-De Fan$^1$
\quad
Chia-Che Chang$^2$
\quad
Yi-Chen Lo$^2$
\\
Yu-Chee Tseng$^1$
\quad
Jiun-Long Huang$^1$
\quad
Yu-Lun Liu$^1$\vspace{0.5em}
\\
\centerline{$^1$National Yang Ming Chiao Tung University \quad $^2$MediaTek Inc.}\vspace{0.5em}
\\
{\url{https://chenwei891213.github.io/GCC/}}
}

\newif\ifreview 
\newif\ifarxiv \newcommand{\arxiv}{\arxivtrue}
\newif\ifcamera 
\newif\ifrebuttal 

\arxiv

\pdfoutput=1
\documentclass[10pt,twocolumn,letterpaper]{article}
\ifreview \usepackage[review]{cvpr} \fi
\ifarxiv \usepackage[pagenumbers]{cvpr} \fi
\ifrebuttal \usepackage[rebuttal]{cvpr} \fi
\ifcamera \usepackage{cvpr} \fi

\usepackage{graphicx}	
\usepackage{amsmath}	
\usepackage{amssymb}	
\usepackage{booktabs}
\usepackage{times}
\usepackage{microtype}
\usepackage{epsfig}
\usepackage{caption}
\usepackage{float}
\usepackage{placeins}
\usepackage{color, colortbl}
\usepackage{stfloats}
\usepackage{enumitem}
\usepackage{tabularx}
\usepackage{xstring}
\usepackage{multirow}
\usepackage{xspace}
\usepackage{url}
\usepackage{subcaption}
\usepackage{xcolor}
\usepackage[hang,flushmargin]{footmisc}
\usepackage[ruled,vlined]{algorithm2e}
\ifcamera \usepackage[accsupp]{axessibility} \fi
\usepackage{gensymb}

% ...

\ifarxiv  \fi

\newcommand{\R}[1]{{%
    \textbf{%
        \ifstrequal{#1}{1}{\textcolor{red}{R#1}}{%
        \ifstrequal{#1}{2}{\textcolor{blue}{R#1}}{%
        \ifstrequal{#1}{3}{\textcolor{magenta}{R#1}}{%
        \ifstrequal{#1}{4}{\textcolor{teal}{R#1}}{%
                           \textcolor{cyan}{R#1}%
        }}}}%
    }%
}}

\usepackage{xr-hyper}

\makeatletter
\newcommand*{\addFileDependency}[1]{
  \typeout{(#1)}
  \@addtofilelist{#1}
  \IfFileExists{#1}{}{\typeout{No file #1.}}
}

\makeatother
\newcommand*{\myexternaldocument}[1]{
    \externaldocument{#1}
    \addFileDependency{#1.tex}
    \addFileDependency{#1.aux}
}

\definecolor{cvprblue}{rgb}{0.21,0.49,0.74}
\usepackage[pagebackref,breaklinks,colorlinks,allcolors=cvprblue]{hyperref}
\usepackage[capitalize]{cleveref}
\crefname{section}{Sec.}{Secs.}
\crefname{table}{Table}{Tables}
\crefname{figure}{Fig.}{Figs.}

\ifarxiv \crefname{appendix}{App.}{Apps.}
\else \crefname{appendix}{Suppl.}{Suppls.} \fi

\unless\ifarxiv \myexternaldocument{_supplementary} \fi

\begin{document}
\title{\paperTitle}
\author{\authorBlock}

\twocolumn[{%
\renewcommand\twocolumn[1][]{#1}%
\maketitle
\begin{center}
\centering
\captionsetup{type=figure}
\vspace{-7mm}
\resizebox{1.0\textwidth}{!} 
{
\includegraphics[width=\textwidth]{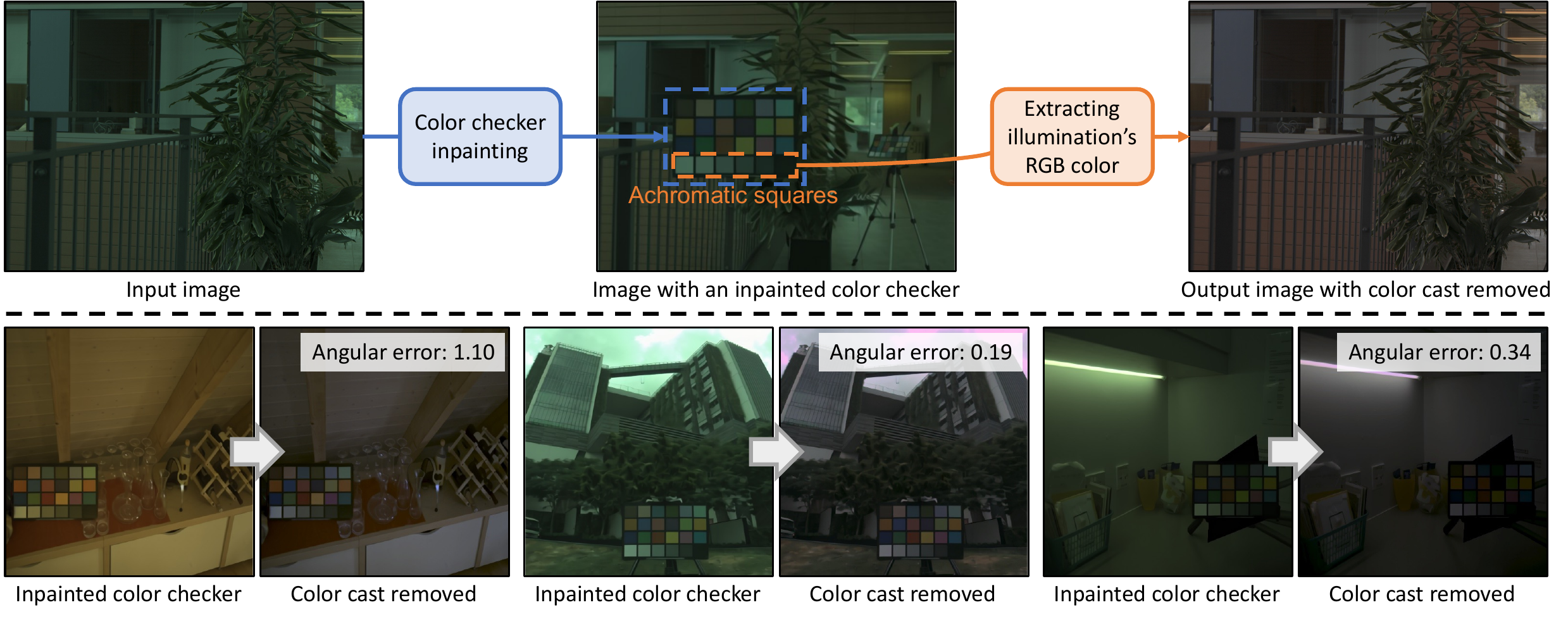}
}
\vspace{-8mm}
\caption{
\textbf{Our method performs color constancy through diffusion-based color checker inpainting.} (\emph{top left}) Given an input image, we first inpaint a color checker with Stable Diffusion, aligning the achromatic (gray) squares to accurately reflect the scene illumination (\emph{top middle}). The RGB color extracted from the achromatic squares is then used to remove the color cast from the input image (\emph{top right}). (\emph{Bottom}) Our approach leverages the strong priors of pre-trained diffusion models to accurately estimate scene illumination without requiring physical color checkers during capture, enabling accurate white balance correction across diverse scenes.
}
\label{fig:teaser}
 \end{center}
}]

\maketitle

\begin{abstract}
\vspace{-5mm}

Color constancy methods often struggle to generalize across different camera sensors due to varying spectral sensitivities. We present GCC, which leverages diffusion models to inpaint color checkers into images for illumination estimation. Our key innovations include (1) a single-step deterministic inference approach that inpaints color checkers reflecting scene illumination, (2) a Laplacian decomposition technique that preserves checker structure while allowing illumination-dependent color adaptation, and (3) a mask-based data augmentation strategy for handling imprecise color checker annotations. By harnessing rich priors from pre-trained diffusion models, GCC demonstrates strong robustness in challenging cross-camera scenarios. These results highlight our method's effective generalization capability across different camera characteristics without requiring sensor-specific training, making it a versatile and practical solution for real-world applications.

\end{abstract}
\vspace{-5mm}
\section{Introduction}
\label{sec:intro}

Color constancy is a crucial aspect of computer vision, focused on determining the illumination of a scene to ensure that colors are accurately represented under varying lighting conditions. This process is essential for maintaining a consistent color appearance and for applications ranging from photography to autonomous driving. Traditional statistics-based methodologies \cite{buchsbaum1980spatial, land1977retinex, forsyth1990novel, van2007edge, joze2012role, qian2019finding, barnard2002comparison, finlayson2004shades} rely on various statistical assumptions about scene color distributions. While these methods are computationally efficient, they often struggle in challenging scenes when their underlying assumptions are violated, especially in environments with multiple illuminants or complex lighting conditions.

In contrast, deep learning-based methods \cite{hu2017fc4, Bianco2015ColorCU, Lou2015ColorCB} have significantly advanced the field of color constancy through their ability to learn complex illumination patterns from training data. These approaches typically employ convolutional neural networks with various architectures to achieve state-of-the-art performance, particularly in challenging illumination scenarios.

However, a challenge in learning-based color constancy is that models are often constrained to specific camera sensors due to variations in spectral sensitivities. Recent cross-camera approaches \cite{igtn, bianco2019quasi, afifi2019sensor, afifi2021cross, lo2021clcc, yu2020cascading} have made strides in addressing this limitation through techniques including metric learning, quasi-unsupervised learning, and device-independent representations. Building upon these advances, we explore an approach that leverages foundation models to enhance cross-camera performance.


Inspired by the recent success of DiffusionLight~\cite{phongthawee2024diffusionlight}, which leverages pre-trained diffusion models for lighting estimation by inpainting a chrome ball, we propose Generative Color Constancy (GCC), a novel approach that harnesses the rich priors of foundation models to overcome the camera-specific limitations of traditional methods. Unlike DiffusionLight~\cite{phongthawee2024diffusionlight}, which focuses on HDR lighting estimation, our method adapts the concept to color constancy by inpainting a color checker into the input image. Color checkers are widely used calibration tools in color science, and our diffusion model generates one with colors that accurately represent the scene's illumination. By analyzing the generated color checker's patches, we can effectively estimate the scene's illuminant. However, diffusion models typically generate outputs stochastically, which is undesirable for color constancy applications requiring consistency. Drawing insights from recent work on deterministic fine-tuning of image-conditional diffusion models~\cite{garcia2024fine}, we design a deterministic pipeline that produces consistent illumination estimates while preserving the powerful generalization capabilities of the underlying foundation model. Our approach eliminates the need for camera-specific training data, achieving robust performance across different camera sensors and scene types.

In summary, we make the following contributions:
\begin{itemize}
    \item We propose a novel color constancy method that leverages diffusion models to inpaint a color checker, which serves as a virtual reference for illumination estimation.
    
    \item We introduce a Laplacian decomposition technique that enhances the model to generate color checkers that maintain structure while adapting to scene illumination, improving color extraction accuracy.
    
    \item We design a deterministic single step inference pipeline that avoids introducing noise during training and inference, resulting in consistent results and improved computational efficiency compared to traditional diffusion processes.
\end{itemize}

\section{Related Work}
\label{sec:related}

\noindent {\bf Color Constancy and White Balance.}
Color constancy research spans statistical-based and learning-based approaches. Statistical methods like Gray World \cite{buchsbaum1980spatial}, Gray Edge \cite{WeijerGG07}, Shades-of-Gray \cite{finlayson2004shades}, Bright Pixels \cite{joze2012role}, and Gray Index \cite{qian2019finding} make assumptions about scene color statistics but struggle with challenging scenes.
Learning-based methods have proven more effective, evolving from gamut mapping \cite{barnard2000improvements,chakrabarti2011color} and regression models \cite{funt2004estimating} to more advanced techniques. Notable developments include CCC \cite{barron2015convolutional} and FFCC \cite{barron2017fast}, which use convolutional processing and frequency-domain optimization. Deep learning approaches like FC4 \cite{hu2017fc4}, DS-Net \cite{shi2016deep}, RCC-Net \cite{8237844}, and C4 \cite{yu2020cascading} further improve performance with various neural network architectures. 

A key challenge is camera-specific spectral sensitivity \cite{afifi2019sensor,gao2017improving}, requiring retraining or calibration for new sensors \cite{liba2019handheld}. Recent solutions include IGTN's \cite{igtn} metric learning, quasi-unsupervised learning \cite{bianco2019quasi}, and cross-dataset approaches \cite{Koskinen2020CrossdatasetCC}. SIIE \cite{afifi2019sensor} proposes sensor-independent illumination estimation, while C5 \cite{afifi2021cross} uses unlabeled target camera images during inference, and CLCC \cite{lo2021clcc} employs contrastive learning to improve feature representations. Our work leverages pre-trained diffusion models for color checker inpainting, utilizing their rich knowledge priors to offer a novel approach to illumination estimation with enhanced generalization capability across different camera sensors.

\vspace{3pt}  \noindent {\bf Image-conditional Diffusion Models.}
Denoising Diffusion Probabilistic Models (DDPMs) \cite{sohl2015deep} achieve state-of-the-art generation by reversing a noising process with UNet architectures \cite{ronneberger2015u}, demonstrating excellence in density estimation and sample quality \cite{kingma2021variational, dhariwal2021diffusion}.
Latent Diffusion Models (LDMs) \cite{rombach2021highresolution} improved efficiency by operating in compressed latent space and introduced cross-attention conditioning. This enabled powerful inpainting capabilities, demonstrated by Blended Diffusion \cite{avrahami2022blendeddiffusion, avrahami2023blendedlatent}, Paint-by-Example \cite{yang2023paint}, ControlNet \cite{zhang2023adding}, and IP-Adapter \cite{ye2023ip-adapter}.
Recent work identified that perceived limitations were often due to DDIM scheduler implementation issues \cite{lin2024common} rather than fundamental constraints. Our work leverages these insights to effectively adapt diffusion models for color checker inpainting in illumination estimation.

\vspace{3pt}  \noindent {\bf Learning-based Lighting Estimation.}
Lighting estimation methods traditionally use physical probes like mirror balls \cite{Debevec1998}, 3D objects \cite{weber2018learning, lombardi2015reflectance}, eyes \cite{nishino2004eyes}, or faces \cite{calian2018faces, yi2018faces}. Early probe-free approaches used limited models like directional lights \cite{karsch2011rendering}, sky models \cite{hosek2012analytic, hold2017outdoor}, or spherical harmonics \cite{garon2019fastspatialvary}.
Modern methods focus on HDR environment maps, pioneered by Gardner et al. \cite{garder2017lavelindoor}. DeepLight \cite{legendre2019deeplight} and EverLight \cite{dastjerdi2023everlight} handle both indoor and outdoor scenes, while StyleLight \cite{wang2022stylelight} uses GANs for joint LDR-HDR prediction. Some works explore panorama outpainting \cite{akimoto2019360outpainting2stategan, dastjerdi2022immersegan} but struggle with HDR \cite{dastjerdi2023everlight}.
Recently, DiffusionLight \cite{phongthawee2024diffusionlight} introduced virtual chrome ball synthesis using diffusion models. Our work follows a similar direction but focuses on color checker inpainting for illumination estimation.

\vspace{3pt}  \noindent {\bf Fine-tuning Strategies for Diffusion Models.}
For personalization, DreamBooth \cite{ruiz2022dreambooth} pioneered special token fine-tuning, while \citeauthor{gal2022image} \cite{gal2022image} and \citeauthor{voynov2023P+} \cite{voynov2023P+} proposed learned word embeddings approaches. Similar to DreamBooth, our method fine-tunes pre-trained diffusion models to bind specific visual characteristics to our target domain, enabling a consistent generation of color checkers that reflect scene illumination.
For efficiency and fine-tuning strategies, LoRA \cite{hu2021lora} introduced low-rank weight changes, while SVDiff \cite{han2023svdiff} and orthogonal fine-tuning \cite{oft2023} proposed alternative parameterizations. For geometry estimation, Marigold \cite{ke2023marigold} demonstrated successful fine-tuning using synthetic data. Inspired by \citeauthor{garcia2024fine} \cite{garcia2024fine}, who showed that simple fine-tuning approaches can be highly effective for deterministic tasks involving low-frequency image components, we adopt their full fine-tuning strategy for our color checker inpainting task. This approach aligns well with our color constancy problem, which primarily focuses on modifying the low-frequency characteristics of color checkers.
\section{Method}

\begin{figure*}[t]
    \centering
    \includegraphics[width=1\linewidth]{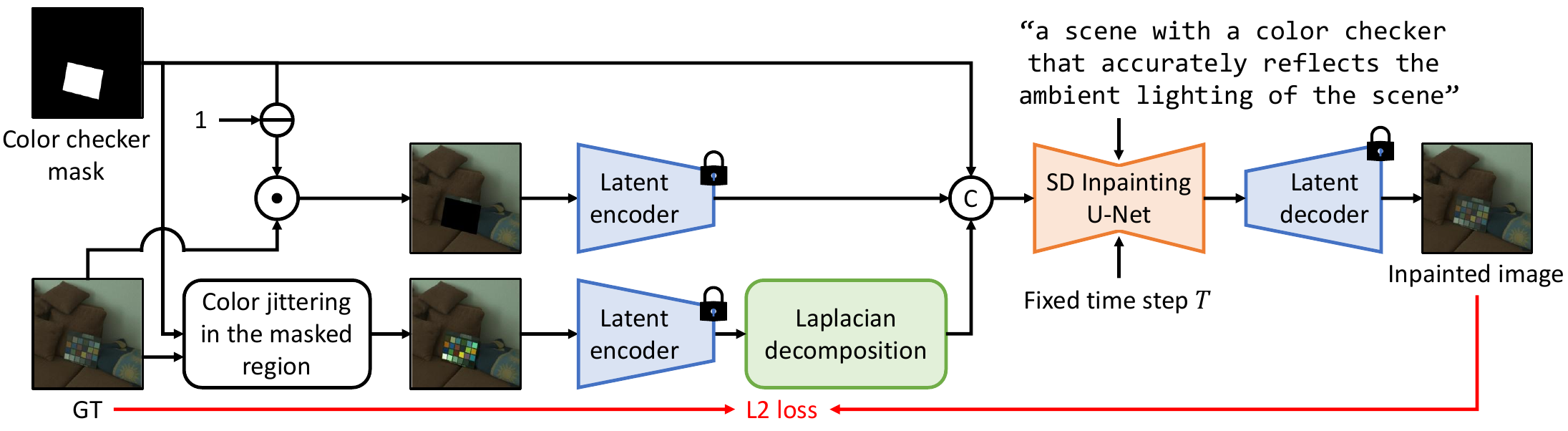}
\vspace{-7mm}
\caption{\textbf{Overview of our training pipeline.} Starting from stable-diffusion-2-inpainting~\cite{rombach2021highresolution}, we enable color checker generation through end-to-end fine-tuning. Given a ground truth color checker image and its mask, we apply color jittering in the masked region. The input image latent passes through Laplacian decomposition before being concatenated with the masked image latent and the resized mask for the SD Inpainting U-Net. The model is trained with an L2 loss between the inpainted output and ground truth image at a fixed timestep $T$.}
    \label{fig:pipeline}
\end{figure*}

\begin{figure*}[t]
    \centering
    \vspace{-1mm}
    \includegraphics[width=1\linewidth]{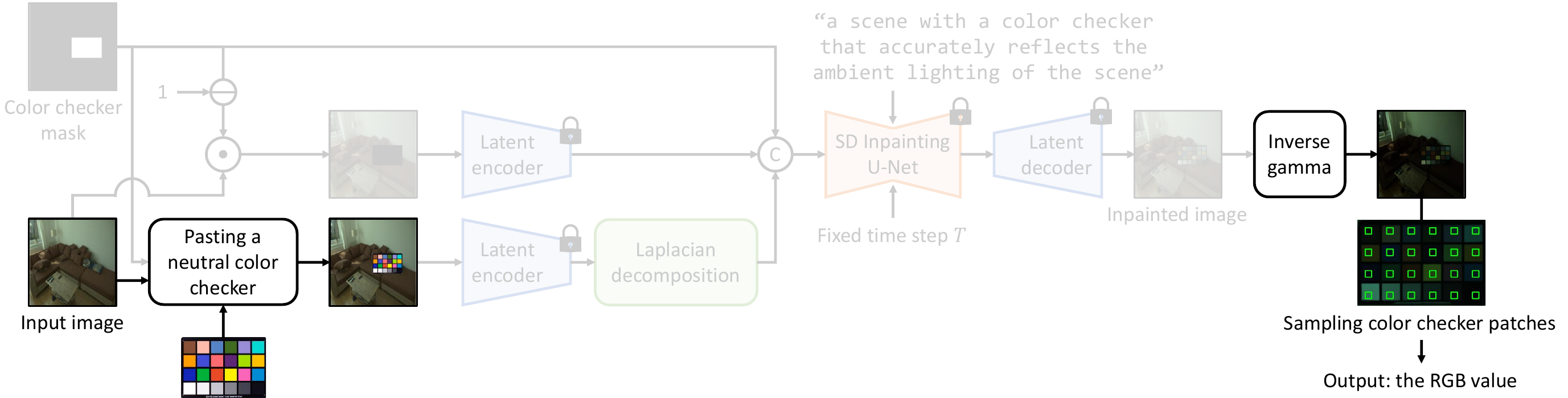}
\vspace{-7mm}
    \caption{\textbf{Overview of our inference pipeline for illumination estimation.} A neutral color checker is pasted onto the input image, which is then encoded into the latent space. The input latent is processed through Laplacian composition before being concatenated with the masked image latent and the resized mask. The modified U-Net generates an inpainted result at fixed timestep $T$. After inverse gamma correction, we sample the color checker patches to obtain the final RGB illumination value. We highlight the steps and components that are different from the training pipeline.}
    \label{fig:inference}
\end{figure*}

\label{sec:method}
Instead of directly predicting environmental RGB light, we propose to leverage diffusion models' rich priors to inpaint a color checker into the scene and extract illumination colors from it. As shown in \cref{fig:pipeline,fig:inference}, our pipeline consists of (1) During training, we fine-tune a diffusion-based inpainting model at timestep t=T with images containing color checkers, optimizing for deterministic single-step inference (Sec. \ref{sec:network_architecture}-\ref{sec:end_to_end_finetune}). (2) We introduce Laplacian decomposition to maintain the checker's high-frequency structure while allowing illumination-aware color adaptation (Sec. \ref{sec:laplacian_composition}). (3) At inference time, we composite a neutral color checker into a given scene and use our fine-tuned model to inpaint it according to the scene illumination, from which we extract the scene's light color information (Sec. \ref{sec:inference}). 

\subsection{Network Architecture}
\label{sec:network_architecture}
We base our model on stable-diffusion-2-inpainting {\cite{rombach2021highresolution}} for its specialized local editing capability. The model consists of a VAE encoder-decoder pair ($\mathcal{E}$, $\mathcal{D}$) and a U-Net denoising backbone. Given an RGB image $I \in \mathbb{R}^{H\times W\times 3}$ and a binary mask $M \in \{0,1\}^{H\times W}$ indicating the color checker region, we first encode both the masked image and the original image into the latent space as $z_{\text{masked}} =  \mathcal{E}(I \odot (1 - M))$ and $z = \mathcal{E}(I)$, where $\odot$ denotes element-wise multiplication. The mask $M$ is downscaled by a factor of 8 to match the latent resolution as $M' \in \mathbb{R}^{h\times w}$, where $h = H/8, w = W/8$. During training, the U-Net denoiser $\epsilon_\theta$ takes as input the concatenation of the noised latent $z_t$, the downscaled mask $M'$, and the masked image latent $z_{\text{masked}}$ along the channel dimension as $z_{\text{combined}} = [z_t, M', z_{\text{masked}}] \in \mathbb{R}^{h\times w\times(2d+1)}$, where $d$ is the latent dimension. Together with the timestep $t$ and text embedding $c$, the denoiser is trained to predict the noise as $\epsilon_\theta(z_{\text{combined}}, t, c) \rightarrow \mathbb{R}^{h\times w\times d}$. At inference time, we obtain the final inpainted result by decoding the denoised latent $\hat{I} = \mathcal{D}(z_0)$, where only the color checker region is modified while leaving the rest unmodified, making this architecture particularly suitable for the color constancy task.

\begin{figure}[t]
    \centering
    \includegraphics[width=1\columnwidth]{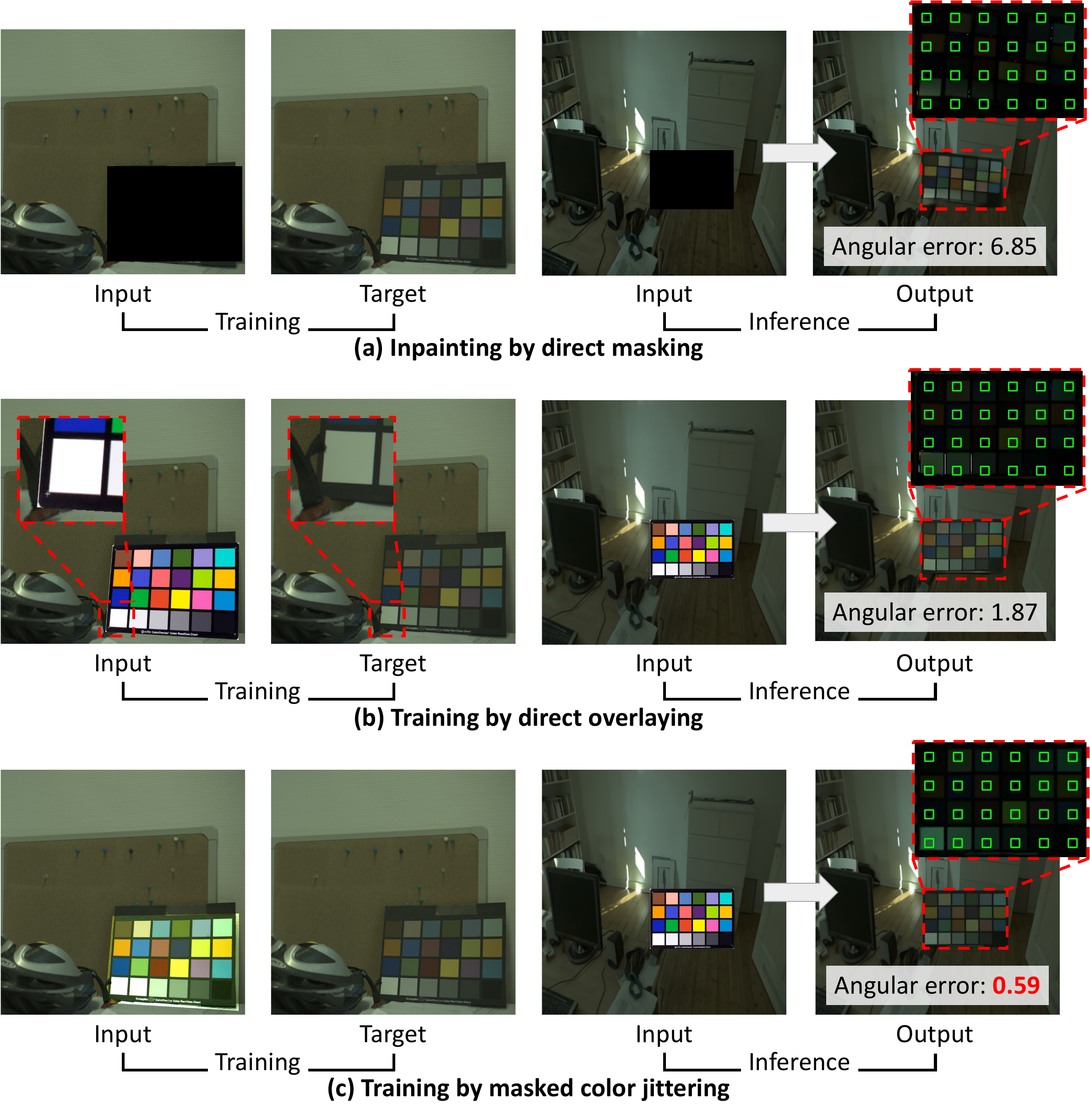}
    \vspace{-8mm}
    \caption{\textbf{Analysis of color checker alignment strategies.} (a) Direct inpainting on masked regions leads to poor color checker structure. This is because we do not provide guidance on the desired color checker structure, causing the model to generate contours that do not meet our expectations. (b) Using a homography transform to overlay a template suffers from pixel-level misalignment due to imprecise bounding box annotations. (c) Our mask color jittering approach overcomes corner annotation limitations by allowing the model to generate geometrically consistent color checker structures while accurately reflecting scene illumination.}
    \label{fig:direct_inpaint}
\end{figure}

\subsection{End-to-End Fine-Tuning}
\label{sec:end_to_end_finetune}

\noindent {\bf Training.}
Although pre-trained diffusion models like SD and SD inpainting \cite{rombach2021highresolution} have been exposed to diverse image collections, additional fine-tuning is crucial for generating precise color checkers that accurately reflect environmental illumination. As shown in experiments \cref{fig:wo_fig}, fine-tuning significantly impacts the model's ability to generate color checkers that faithfully represent scene illumination.


Although SDEdit~\cite{meng2022sdeditguidedimagesynthesis} could be applied to our task, it faces a fundamental trade-off in noise level selection. On one hand, insufficient noise fails to effectively suppress the original chromatic information from the input image, making it difficult to adapt to the target scene illumination. On the other hand, excessive noise, while better at removing unwanted color information, can disrupt the structural consistency between the generated result and the input reference. Furthermore, for color constancy tasks, maintaining a one-to-one correspondence between input and output is essential. While traditional diffusion models' stochastic nature allows for ensemble improvements through multiple inferences, this comes at an increased computational cost.


Following \cite{garcia2024fine}, we adopt an end-to-end fine-tuning approach that enables single-step deterministic inference while maintaining high-quality color checker generation. Specifically, we fine-tune the inpainting U-Net at a fixed timestep $t = T$ as shown in \cref{fig:pipeline}. 
{Given an input image $I$ and its corresponding mask $\mathbf{M}$, we first obtain the augmented image $I_{\text{aug}}$ by applying color jittering to the masked region. We then obtain its latent representation through the VAE encoder, $\mathbf{z}^* = \mathcal{E}(I_{\text{aug}})$. The latent representation is processed through Laplacian decomposition to extract high-frequency components, $\mathbf{z}_h = \mathcal{H}(\mathbf{z}^*)$. For single-step prediction, we directly set the noise term $\boldsymbol{\epsilon} = \mathbf{0}$ in the forward process: $\mathbf{z}_T = \sqrt{\bar{\alpha}_T}\mathbf{z}_h + \sqrt{1-\bar{\alpha}_T}\boldsymbol{\epsilon}$. The denoised latent is then obtained through $\hat{\mathbf{z}}_0 = \sqrt{\bar{\alpha}_T}\mathbf{z}_T - \sqrt{1-\bar{\alpha}_T}\epsilon_\theta(\mathbf{z}_{\text{combined}}, T, c)$, where $\mathbf{z}_{\text{combined}} = [\mathbf{z}_T, \mathbf{M}', \mathbf{z}_{\text{masked}}] \in \mathbb{R}^{h \times w \times (2d+1)}$ represents the concatenated input features along the channel dimension, and $c$ denotes the text condition. Finally, we decode the latent to obtain the inpainted image: $\hat{I} = \mathcal{D}(\hat{\mathbf{z}}_0)$. The model is optimized using a mean squared error loss:
%
\begin{equation}
\mathcal{L} = \frac{1}{HW}\sum_{i,j} (I^*_{i,j} - \hat{I}_{i,j})^2, 
\end{equation}
where $(i,j)$ denotes the pixel coordinates, and $H$ and $W$ are the height and width of the image, respectively.

\vspace{3pt}  \noindent {\bf Color checker misalignment issue.}
Existing color constancy datasets \cite{Cheng:14,4587765} only provide rough bounding boxes for color checkers instead of precise corner point locations. This hinders our ability to accurately align the standard sRGB color checker with the one in the original image, affecting the model's learning of the transformation from standard to harmonized colors. To overcome this limitation, we designed a mask region-based data augmentation method.

We first analyze two intuitive solutions: directly masking and allowing the model to perform inpainting. This approach results in generated color checkers with contours that do not meet our expectations, making accurate color extraction from the patches difficult (\cref{fig:direct_inpaint} (a)). The second solution involved overlaying the color checker template directly onto the original image (\cref{fig:direct_inpaint} (b)). However, due to the absence of precise corner point locations, alignment with the raw checker remains imperfect at a per-pixel level even when using homography transform.

\vspace{3pt}  \noindent {\bf Masked color jittering.}
Therefore, we further explored a third approach: directly applying strong color jittering to the mask region (\cref{fig:direct_inpaint} (c)). This seemingly counterintuitive method aims to destroy clues that may leak sensor-specific information, forcing the model to rely on information outside the mask region to reconstruct the original color checker that aligns with the ground truth. 

Random color jittering on masked checkers helps our model learn robust mappings between neutral color references and scene-specific lighting appearances. The augmented image $I_{\text{aug}}$ is obtained by:
\begin{small}
\begin{equation}
I_{\text{aug}} = (1-M) \odot I + M \odot \mathcal{T}(I),
\end{equation}
\end{small}
where $I$ is the input image, $M$ is the binary mask, $\odot$ denotes element-wise multiplication, and $\mathcal{T}(\cdot)$ represents the color jittering function that randomly applies brightness, contrast, and saturation adjustments to the masked region. By randomly perturbing the color checker region, we force the model to rely on contextual illumination cues rather than local color checker patterns. This approach overcomes the limitations of imprecise annotations in existing datasets and enhances the model's ability to learn accurate illumination estimation from scene context.

\subsection{Laplacian Decomposition}
\label{sec:laplacian_composition}
Although mask color jittering addresses the imprecise corner annotation issue, the randomness in jittering may occasionally allow low-frequency information leakage from the masked region. This could cause the model to simply \textit{reconstruct} the masked area rather than \textit{harmonize} it with the scene illumination. To address this issue, we introduce the Laplacian decomposition technique.

By extracting only the high-frequency components of the input image through Laplacian decomposition, our approach serves two purposes: First, it preserves the structural details needed to generate a color checker that faithfully maintains the patch layout of our pre-pasted reference. Second, it minimizes the influence of low-frequency color information, encouraging the model to focus on harmonizing the generated color checker with the scene illumination rather than reconstructing the original colors.
%
The key benefit of Laplacian decomposition, as shown in \cref{fig:wo_fig}, allows the model to generate color checkers that maintain structural consistency while correctly reflecting scene illumination, enabling accurate illumination estimation.

\subsection{Inference}
\label{sec:inference}
The complete inference pipeline of our method is illustrated in \cref{fig:inference}, which consists of the following steps:

\vspace{3pt}  \noindent {\bf Color checker generation.}
We first composite a fixed-size neutral color checker centered at the mask region. The input image is then gamma-corrected with $\gamma = 2.2$ to transform it to the sRGB domain. This preprocessed image is processed through our model in a single forward pass with fixed timestep $t = T$. The output is then inverse gamma-corrected to obtain the raw domain result.

\vspace{3pt} \noindent {\bf Illumination estimation.} Since we have precise control over initial checker placement and Laplacian decomposition ensures structural preservation, we can reliably extract color information from each patch. Specifically, we directly map the generated checker to a standardized grid, followed by applying fixed masks to sample colors from each patch. The scene illumination is then estimated from the achromatic patches of the color checker.

\section{Experiments}
\label{sec:exp}
\subsection{Experimental Setup}

\noindent {\bf Dataset.}
We use two publicly available color constancy benchmark datasets in our experiments: the NUS-8 dataset~\cite{cheng2014illuminant} and the re-processed Color Checker dataset~\cite{4587765} (referred to as the Gehler dataset). The Gehler dataset~\cite{4587765} contains 568 original images captured by two different cameras, while the NUS-8 dataset~\cite{cheng2014illuminant} contains 1736 original images captured by eight different cameras. Each image in both datasets includes a Macbeth Color Checker chart, which serves as a reference for the ground-truth illuminant color.

\vspace{3pt}  \noindent {\bf Evaluation metrics.}
To evaluate the performance of color constancy methods, we use the standard angular error metric, which measures the angular difference between the estimated illuminant and the ground-truth illuminant. Specifically, the angular error $\theta$ between an estimated illuminant vector $\hat{\mathbf{y}}$ and the ground-truth illuminant vector $\mathbf{y}$ is defined as:
\begin{small}
\begin{equation}
\theta = \arccos\left(\frac{\hat{\mathbf{y}} \cdot \mathbf{y}}{|\hat{\mathbf{y}}| |\mathbf{y}|}\right)
\end{equation}
\end{small}
The angular error is measured in degrees, with smaller values indicating better estimation accuracy. Following previous works, we report the following statistics of the angular error.

\subsection{Implementation Details}
Our implementation is based on the stable-diffusion-2-inpainting model~\cite{rombach2021highresolution} using PyTorch. All input images are resized to 512$\times$512 resolution for both training and inference. Since the pre-trained VAE was trained on sRGB images, we apply a gamma correction of $\gamma = 1/2.2$ on linear RGB images before encoding to minimize the domain gap. Conversely, after VAE decoding, we apply inverse gamma correction to convert the output back to the linear domain for metric evaluation.

Following parameter settings from~\cite{garcia2024fine}, we train our models using the Adam optimizer with an initial learning rate of $5 \times 10^{-5}$ and apply an exponential learning rate decay after a 150-step warm-up period. For cross-dataset evaluation, when training on the Gehler dataset and testing on NUS-8, we use a batch size of 8 with no gradient accumulation for 20k iterations. When training on NUS-8 and testing on the Gehler dataset, we use a batch size of 8 with gradient accumulation over 2 steps (effective batch size of 16) for 20k iterations. 

For data augmentation, we follow FC4~\cite{hu2017fc4} to rescale images by random RGB values in [0.6, 1.4] in the raw domain, noting that we only rescale input images since our training does not require ground truth illumination. We also apply mask color jittering to handle imprecise color checker annotations. For Laplacian decomposition, we use a two-level pyramid ($L=2$) to balance the preservation of high-frequency structural details and the suppression of low-frequency color information. All experiments were conducted on an NVIDIA RTX 4090 GPU. Additional implementation details are provided in the supplementary material.

\begin{table*}[t]
\caption{\textbf{Camera-agnostic evaluation.} All results are in units of degrees.}
\label{tab:camera_agnostic}
\vspace{-3mm}
\small  
\centering
\resizebox{0.9\textwidth}{!}{  
\begin{tabular}{l|ccccc|ccccc}
\toprule
\multicolumn{1}{r|}{Training set $\rightarrow$ Testing set} & \multicolumn{5}{c|}{NUS-8 dataset~\cite{cheng2014illuminant} $\rightarrow$ Gehler dataset~\cite{4587765}} & \multicolumn{5}{c}{Gehler dataset~\cite{4587765}
$\rightarrow$ NUS-8 dataset~\cite{cheng2014illuminant}} \\ \cmidrule(lr){2-6} \cmidrule(lr){7-11}
Method & Mean & Median & Tri-mean & Best 25\% & Worst 25\% & Mean & Median & Tri-mean & Best 25\% & Worst 25\% \\
\midrule
\multicolumn{11}{l}{\textbf{Statistical Methods}} \\
White-Path~\cite{brainard1986analysis} & 7.55 & 5.68 & 6.35 & 1.45 & 16.12 & 9.91 & 7.44 & 8.78 & 1.44 & 21.27 \\
Gray-World~\cite{buchsbaum1980spatial} & 6.36 & 6.28 & 6.28 & 2.33 & 10.58 & 4.59 & 3.46 & 3.81 & 1.16 & 9.85 \\
1st-order Gray-Edge~\cite{van2007edge} & 5.33 & 4.52 & 4.73 & 1.86 & 10.43 & 3.35 & 2.58 & 2.76 & 0.79 & 7.18 \\
2nd-order Gray-Edge~\cite{van2007edge} & 5.13 & 4.44 & 4.62 & 2.11 & 9.26 & 3.36 & 2.70 & 2.80 & 0.89 & 7.14 \\
Shades-of-Gray~\cite{finlayson2004shades} & 4.93 & 4.01 & 4.23 & 1.14 & 10.20 & 3.67 & 2.94 & 3.03 & 0.99 & 7.75 \\
General Gray-World~\cite{barnard2002comparison} & 4.66 & 3.48 & 3.81 & 1.00 & 10.09 & 3.20 & 2.56 & 2.68 & 0.85 & 6.68 \\
Grey Pixel (edge)~\cite{yang2015efficient} & 4.60 & 3.10 & - & - & - & 3.15 & 2.20 & - & - & - \\
\citeauthor{cheng2014illuminant}~\cite{cheng2014illuminant} & 3.52 & 2.14 & 2.47 & \cellcolor{orange!25}0.50 & 8.74 & 2.92 & 2.04 & 2.24 & \cellcolor{orange!25}0.62 & 6.61 \\
LSRS~\cite{gao2014efficient} & 3.31 & 2.80 & 2.87 & 1.14 & 6.39 & 3.45 & 2.51 & 2.70 & 0.98 & 7.32 \\
GI~\cite{qian2019finding} & 3.07 & \cellcolor{red!25}1.87 & \cellcolor{orange!25}2.16 & \cellcolor{red!25}0.43 & 7.62 & 2.91 & \cellcolor{orange!25}1.97 & 2.13 & \cellcolor{red!25}0.56 & 6.67 \\
\midrule
\multicolumn{11}{l}{\textbf{Learning-based Methods}} \\
Bayesian~\cite{gehler2008bayesian} & 4.75 & 3.11 & 3.50 & 1.04 & 11.28 & 3.65 & 3.08 & 3.16 & 1.03 & 7.33 \\
\citeauthor{chakrabarti2015color}~\cite{chakrabarti2015color} & 3.52 & 2.71 & 2.80 & 0.86 & 7.72 & 3.89 & 3.10 & 3.26 & 1.17 & 7.95 \\
FFCC~\cite{barron2017fast} & 3.91 & 3.15 & 3.34 & 1.22 & 7.94 & 3.19 & 2.33 & 2.52 & 0.84 & 7.01 \\

SqueezeNet-FC$^4$~\cite{hu2017fc4} & \cellcolor{yellow!25}3.02 & 2.36 & 2.50 & 0.81 & \cellcolor{yellow!25}6.36 & \cellcolor{yellow!25}2.40 & 2.03 & \cellcolor{orange!25}2.10 & 0.70 & \cellcolor{yellow!25}4.80 \\
C$^4_{\text{SqueezeNet-FC4}}$~\cite{yu2020cascading}  & \cellcolor{orange!25}2.73 & 
 \cellcolor{yellow!25}2.20 & \cellcolor{yellow!25}2.28 & \cellcolor{yellow!25}0.72 & \cellcolor{orange!25}5.69 & \cellcolor{red!25}2.28 & \cellcolor{red!25}1.90 & \cellcolor{red!25}1.97 & 0.67 & \cellcolor{orange!25}4.60 \\
SIIE~\cite{afifi2019sensor} & 3.72 & 2.46 & 2.79 & 1.02 & 8.51 & 4.24 & 3.88 & 3.93 & 1.45 & 7.66 \\
CLCC~\cite{lo2021clcc} &{3.05} & {2.44} & {2.51} & {0.89} & {6.30} & 3.42 & 2.95 & 3.06 & 0.94 & 6.70 \\
C$^5$~\cite{afifi2021cross} &3.34  &2.57  &2.68  & 0.78  &7.39  &2.65  & \cellcolor{yellow!25}1.98  &2.14  & \cellcolor{yellow!25}0.66  &5.72  \\

Ours& \cellcolor{red!25}2.35  & \cellcolor{orange!25}2.02  & \cellcolor{red!25}2.06  &0.78
&\cellcolor{red!25}4.57  & \cellcolor{orange!25}2.38 &{2.01}  &\cellcolor{yellow!25}{2.10}  &{0.80}  &\cellcolor{red!25}4.58 \\


\bottomrule
\end{tabular}
}
\end{table*}

\begin{table}[t]
    \vspace{-1mm}
\caption{\textbf{Leave-one-out evaluation on the NUS-8 Dataset~\cite{cheng2014illuminant}.}}
\label{tab:nus8_leave_one_out}
\vspace{-3mm}
\centering
\resizebox{\columnwidth}{!}{ 
\begin{tabular}{l|ccccc}
\toprule
\textbf{NUS-8 Dataset~\cite{cheng2014illuminant}} & {Mean} & {Med.}  & {Tri.} &{Best 25\%} & {Worst 25\%}\\
\midrule
Gray-world~\cite{buchsbaum1980spatial} & 4.59 & 3.46  & 3.81 & 1.16 & 9.85 \\
Shades-of-Gray~\cite{finlayson2004shades} & 3.67 & 2.94 & 3.03 & 0.98 & 7.75 \\
Local Surface Reflectance~\cite{gao2014efficient} & 3.45 & 2.51 & 2.70  & 0.98 & 7.32\\
PCA-based B/W Colors~\cite{Cheng:14} & 2.93 & 2.33 & 2.42  & 0.78 & 6.13 \\
Grayness Index~\cite{qian2019finding} & 2.91 & 1.97 & \cellcolor{yellow!25}2.13  & \cellcolor{orange!25}0.56 & 6.67 \\
Cross-dataset CC~\cite{koskinen122020cross} & 3.08 & 2.24 & - & - & - \\
Quasi-Unsupervised CC~\cite{bianco2019quasi} & 3.00 & 2.25 & - & - & - \\
SIIE\cite{afifi2019sensor} & \cellcolor{orange!25}2.05 & \cellcolor{red!25}1.50 & - & \cellcolor{red!25}0.52 & \cellcolor{orange!25}4.48 \\
FFCC~\cite{barron2017fast} & 2.87 & 2.14 & 2.30 & 0.71 & 6.23  \\
C5~\cite{afifi2021cross} & \cellcolor{yellow!25}2.54 & \cellcolor{yellow!25}1.90  & \cellcolor{orange!25}2.02 & \cellcolor{yellow!25}0.61 & \cellcolor{yellow!25}5.61\\
Ours & \cellcolor{red!25}2.03 & \cellcolor{orange!25}1.78  & \cellcolor{red!25}1.83 & 0.77 & \cellcolor{red!25}3.69 \\
\bottomrule
\end{tabular}
}

\end{table}

\begin{table}[t]
    \vspace{-1mm}
\caption{\textbf{Leave-one-out evaluation on the Gehler Dataset~\cite{4587765}.}}
\label{tab:gehler_leave_one_out}
\vspace{-3mm}
\resizebox{\columnwidth}{!}{ 
\begin{tabular}{l|ccccc}
\toprule
\textbf{Gehler dataset~\cite{4587765}} & {Mean} & {Med.} & {Tri.} & {Best 25\%} & {Worst 25\%} \\
\midrule
Shades-of-Gray~\cite{finlayson2004shades} & 4.93 & 4.01 & 4.23 & 1.14 & 10.20 \\
PCA-based B/W Colors~\cite{Cheng:14} & 3.52 & 2.14 & 2.47 & \cellcolor{yellow!25}0.50 & 8.74 \\
ASM~\cite{8039202} & 3.80 & 2.40 & 2.70 & - & - \\
Woo et al.~\cite{8226796} & 4.30 & 2.86 & 3.31 & 0.71 & 10.14 \\
Grayness Index~\cite{qian2019finding} & 3.07 & \cellcolor{red!25}1.87 & \cellcolor{orange!25}2.16 & \cellcolor{red!25}0.43 & 7.62 \\
Cross-dataset CC~\cite{koskinen122020cross} & 2.87 & 2.21 & - & - & - \\
Quasi-Unsupervised CC~\cite{bianco2019quasi} & 3.46 & 2.23 & - & - & - \\
SIIE~\cite{afifi2019sensor} & \cellcolor{orange!25}2.77 & \cellcolor{orange!25}1.93 & - & 0.55 &  \cellcolor{yellow!25}6.53 \\
FFCC~\cite{barron2017fast} & 2.95 & 2.19 & \cellcolor{yellow!25}2.35 & 0.57 & 6.75 \\

C5~\cite{afifi2021cross} & \cellcolor{red!25}2.50 & \cellcolor{yellow!25}1.99 & \cellcolor{red!25}2.03 & \cellcolor{orange!25}0.47 & \cellcolor{orange!25}5.46 \\
Ours & \cellcolor{yellow!25}2.80 & 2.50 & 2.58 & 1.10 & \cellcolor{red!25}5.00\\
\bottomrule
\end{tabular}
}

\end{table}
\subsection{Results and Comparisons}

\noindent {\bf Evaluation protocols.}
We conduct experiments under three different protocols to comprehensively assess our method's performance and generalization capabilities. 
First, following the camera-agnostic evaluation protocol from C4 \cite{yu2020cascading}, we evaluate robustness against camera sensitivity variations by training on one dataset and testing on another. Specifically, we train on the NUS-8 dataset and test on the Gehler dataset and vice versa. As shown in Table~\ref{tab:camera_agnostic}, our method achieves competitive performance compared to state-of-the-art approaches.
Second, we adopt the leave-one-out protocol from SIIE \cite{afifi2019sensor} to assess performance on unseen camera sensors by excluding images from one camera during training and testing on them. This process is repeated for all cameras, with results in \cref{tab:nus8_leave_one_out} and \cref{tab:gehler_leave_one_out} demonstrating our method's effectiveness. Both protocols highlight that our approach leverages diffusion priors to learn sensor-independent illumination features without requiring camera-specific retraining.
Additionally, we conducted standard three-fold cross-validation on both the NUS-8~\cite{cheng2014illuminant} and Gehler datasets~\cite{4587765}
. As shown in \cref{tab:cross_validation_nus8} and \cref{tab:cross_validation_gehler}, our method achieves performance comparable to other approaches, particularly in worst-case scenarios.

\vspace{3pt}  \noindent {\bf Position-aware Sampling and Consistency.}
\cref{fig:diif_region_inpaint} demonstrates our method's robustness in single-illumination scenes. Unlike prior approaches, our ability to sample at different positions and generate result ensembles enables the quantification of model consistency, showcasing our approach's precision and reliability.

\vspace{3pt}  \noindent {\bf Spatially Varying Illumination in Multi-source Scenes.}
Traditional color constancy methods typically assume a single global illuminant, limiting their applicability in complex lighting scenarios. Our method naturally extends to spatially varying illumination conditions. We evaluated this capability on the LSMI dataset~\cite{kim2021large}, which features challenging multi-illuminant scenes. By dividing each image into a 4$\times$4 grid, inpainting color checkers in each cell, and interpolating these local estimates, our method effectively models different lighting regions. Results in ~\cref{tab:LSMI} demonstrate that our approach can handle complex lighting environments without requiring specific fine-tuning for multi-illuminant data. \cref{fig:spaital_varying} visually confirms our method's ability to adapt to lighting transitions in real-world environments.

\vspace{3pt}  \noindent {\bf Computational Efficiency.}
Our method maintains efficient inference times due to its single-step design. Using an NVIDIA RTX 4090 GPU, it processes a 512$\times$512 image in 180ms, significantly faster than traditional diffusion methods requiring multiple denoising steps as shown in ~\cref{tab:model-comparison}, while preserving accuracy benefits from diffusion priors.

\begin{figure*}[t]
    \centering
    \includegraphics[width=\textwidth]{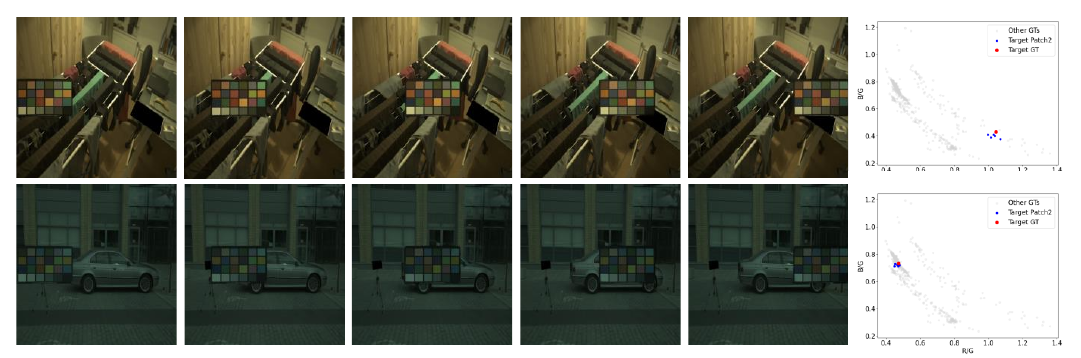}
    \vspace{-8mm}
    \caption{\textbf{Sensitivity to color checker placement.} This figure demonstrates the robustness of our method across various color checker positions under a single light source scenario. The left part displays different placements of color checkers and their corresponding processed results, showing that our method remains effective under challenging warm color temperatures (regions with lower data distribution). The scatter plots on the right quantitatively validate this observation, where the estimated illumination values consistently cluster near the ground truth target, confirming the precision and consistency of our approach.}

    \label{fig:diif_region_inpaint}
    \vspace{-1mm}
\end{figure*}
\begin{table}
    \centering
    \small
    \caption{\textbf{Three-fold cross-validation on NUS-8 dataset~\cite{cheng2014illuminant}.}}
    \label{tab:cross_validation_nus8}
    \vspace{-3mm}
    \resizebox{\columnwidth}{!}{
    \begin{tabular}{l|cccccc}
        \toprule
        NUS-8 dataset~\cite{cheng2014illuminant}  & Mean & Med. & Tri. & Best 25\% & Worst 25\%\\
        \midrule
        CCC \cite{barron2015convolutional} & 2.38 & 1.48 & 1.69 & 0.45 & 5.85  \\
        AlexNet-FC4 \cite{hu2017fc4} & 2.12 & 1.53 & 1.67 & 0.48 & 4.78 \\
        FFCC \cite{barron2017fast}& 1.99 & \cellcolor{red!25}1.31 & \cellcolor{orange!25}1.43 & \cellcolor{red!25}0.35 & 4.75 \\
        C$^4_{\text{SqueezeNet-FC4}}$~\cite{yu2020cascading} & \cellcolor{orange!25}1.96 & \cellcolor{orange!25}1.42 & 1.53 & 0.48 & 4.40 \\
        CLCC \cite{lo2021clcc} & \cellcolor{red!25}1.84 & \cellcolor{red!25}1.31 & \cellcolor{red!25}1.42 & \cellcolor{orange!25}0.41 & \cellcolor{red!25}4.20 \\
        Ours &2.10 & 1.52 & 1.69 & 0.56 & \cellcolor{orange!25}4.38 \\
        \bottomrule
    \end{tabular}
    }
\end{table}

\begin{table}
    \centering
    \small
    \vspace{-1mm}
    \caption{\textbf{Three-fold cross-validation on Gehler dataset~\cite{4587765}.}}
    \label{tab:cross_validation_gehler}
    \vspace{-3mm}
    \resizebox{\columnwidth}{!}{
    \begin{tabular}{l|ccccc}
        \toprule
        Gehler dataset ~\cite{4587765}  & Mean & Med. & Tri. & Best 25\% & Worst 25\%\\
        \midrule
        CCC \cite{barron2015convolutional}& 1.95 & 1.22 & 1.38 & 0.35 & 4.76  \\
        SqueezeNet-FC4 \cite{hu2017fc4} & 1.65 & 1.18 & 1.27 & 0.38 & 3.78  \\
        FFCC \cite{barron2017fast} & 1.61 & \cellcolor{red!25}{0.86} & \cellcolor{red!25}1.02 & \cellcolor{red!25}{0.23} & 4.27  \\
        C$^4_{\text{SqueezeNet-FC4}}$~\cite{yu2020cascading} & \cellcolor{red!25}{1.35} & \cellcolor{orange!25}0.88 & \cellcolor{red!25}{0.99} & 0.28 & \cellcolor{red!25}{3.21} \\
        CLCC\cite{lo2021clcc}  &  \cellcolor{orange!25}1.44 & 0.92 & 1.04 & \cellcolor{orange!25}0.27 & 3.48  \\
        Ours & 1.91 & 1.80 & 1.84 & 0.60 &  \cellcolor{orange!25}{3.46}\\
        \bottomrule
    \end{tabular}
    }
\end{table}


\begin{table}[t]
    \vspace{-1mm}
\caption{\textbf{Zero-shot evaluation on the LSMI Dataset~\cite{kim2021large}.} Mean angular error (MAE) for the spatially varying illumination map.}
\label{tab:LSMI}
\centering
\small
\vspace{-3mm}
\resizebox{\columnwidth}{!}{
\begin{tabular}{l|ccc|ccc}
\toprule
& \multicolumn{3}{c|}{Galaxy} & \multicolumn{3}{c}{Nikon} \\ \cmidrule(lr){2-4} \cmidrule(lr){5-7}
Method & Single & Multi & Mixed & Single & Multi & Mixed \\ 
\midrule
LSMI-H~\cite{kim2021large} & \cellcolor{orange!25}{2.85} & \cellcolor{orange!25}{3.13} & 3.06 & 2.76 & \cellcolor{orange!25}{3.2} & 2.99 \\
LSMI-U~\cite{kim2021large} & 2.95 & \cellcolor{red!25}{2.35} & \cellcolor{red!25}{2.63} & \cellcolor{red!25}{1.51} & \cellcolor{red!25}{2.36} & \cellcolor{red!25}{1.95} \\
Ours & \cellcolor{red!25}{2.05} & 3.44 & \cellcolor{orange!25}{2.82} & \cellcolor{orange!25}{2.10} & 3.58 & \cellcolor{orange!25}{2.88} \\
\bottomrule
\end{tabular}
}
\end{table}

\begin{figure}[ht]
    \centering
    \vspace{-2mm}
    \includegraphics[width=1\linewidth]{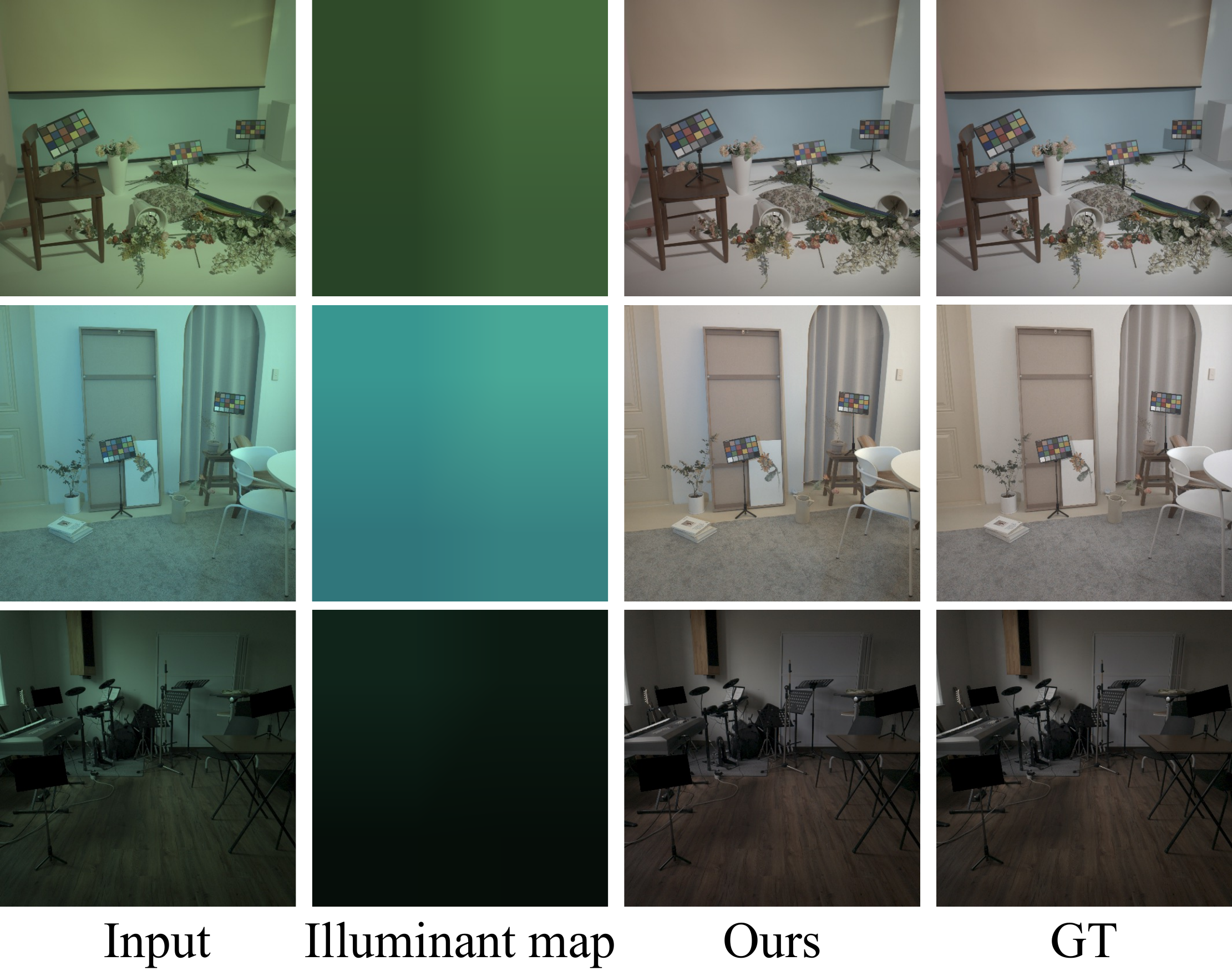}
\vspace{-7mm}
    \caption{\textbf{Spatially varying illumination in multi-source scenes.} From left to right: input image with mixed illumination, illuminant coefficient map showing per-pixel light distribution, our white balanced result, and ground truth white balanced image.}
    \label{fig:spaital_varying}
\end{figure}

\subsection{Ablation Studies}
We conducted a series of ablation experiments to validate the importance of key design choices, including using Laplacian decomposition, noise prediction-based LoRA fine-tuning, and mask-based data augmentation in \cref{table:exp_ablation}.

\begin{table*}[t]
\centering
\small
\caption{\textbf{Comparison between fine-tuned SDXL inpainting and our one-step model.} All metrics are reported in degrees, and inference time is measured on a single 512$\times$512 image using an NVIDIA RTX 4090 GPU. All models are trained on the NUS-8 dataset \cite{cheng2014illuminant} and evaluated on the Gehler dataset \cite{4587765}.}
\vspace{-3mm}
\label{tab:model-comparison}
\begin{tabular}{lccc|cccc}
\toprule
Method & Steps & Ensemble & Inference time (s) & Mean & Median & Best-25\% & Worst-25\% \\
\midrule
SDXL Inpainting (SDEdit) & 25 & 10 & 17.98 & 4.47 & 3.25 & 1.07 & 10.01\\
Full Model& 1 & 1 & \textbf{0.18} & \textbf{2.35} & \textbf{2.02} & \textbf{0.78} & \textbf{4.57} \\
\bottomrule
\end{tabular}
\end{table*}

\begin{table}[t]
\centering
\vspace{-1mm}
\caption{\textbf{Ablation study on key components of our method.} We evaluate the impact of components: Laplacian decomposition (Lap.), color checker inpainting vs. RGB prediction, and masked color jittering (Mask DA). All models are trained on the NUS-8 dataset~\cite{cheng2014illuminant} and evaluated on the Gehler dataset~\cite{4587765}. The results show that our color checker inpainting approach outperforms direct RGB prediction, and the combination with other components (Laplacian decomposition and masked color jittering) yields the best performance. All error metrics are reported in degrees, with lower values indicating better performance.}
\vspace{-3mm}
\label{table:exp_ablation}
\resizebox{\columnwidth}{!}{
\begin{tabular}{cccc|cccc}
\toprule
Noise & Lap. & Inpaint & Mask DA & Mean & Median & Best-25\% & Worst-25\% \\
\midrule
Zeros & - & \checkmark & \checkmark & 3.71 & 2.86 & 1.31 & 7.68 \\
Zeros & \checkmark & \checkmark & - & 3.52 & 2.76 & 1.25 & 6.78 \\
Zeros & - & - & - &2.98 & 2.53 & 1.26 & 6.14 \\
Zeros & \checkmark & \checkmark & \checkmark & \textbf{2.35} & \textbf{2.02} & \textbf{0.78} & \textbf{4.57} \\
\bottomrule
\end{tabular}
}
\vspace{-1mm}
\end{table}

\begin{figure}[t]
    \centering
    \vspace{-2mm}
    \includegraphics[width=1\linewidth]{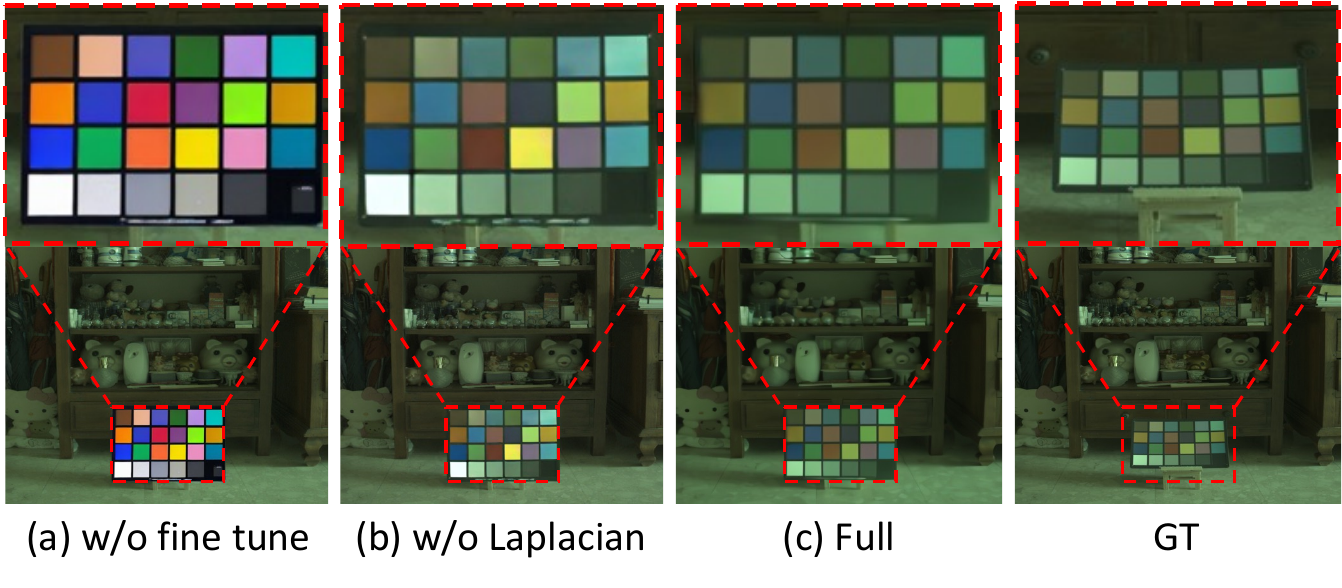}
    \vspace{-8mm}
    \caption{\textbf{Effect of fine-tuning and Laplacian decomposition.} 
    (a) Results without fine-tuning show poor color checker quality due to the domain gap between the pre-trained diffusion model's training data (sRGB images) and our gamma-corrected raw images, leading to disharmonious inpainting results. (b) Results without Laplacian decomposition are biased by low-frequency information from the neutral color checker, leading to inharmonious generation. (c) Our full method with both components produces well-harmonized color checkers that accurately reflect scene illumination.}
    \label{fig:wo_fig}
    \vspace{-1mm}
\end{figure}

\vspace{3pt}  \noindent {\bf Without Laplacian decomposition.}
Without Laplacian decomposition, we use only the VAE encoder's latent representation as input. As shown in ~\cref{fig:wo_fig}, the generated checker is contaminated by low-frequency information from the initial neutral reference, producing disharmonious colors that prevent accurate environmental color estimation.

\vspace{3pt}  \noindent {\bf With noise.}
In this experiment, we used LoRA \cite{hu2021lora} to fine-tune the SDXL inpainting model \cite{rombach2021highresolution} and obtained the final output through ensemble averaging across multiple samples. As shown in ~\cref{tab:model-comparison}, this approach underperforms our final method due to the fundamental trade-off between preserving the color checker's geometry and suppressing low-frequency information from the neutral reference checker.

\vspace{3pt}  \noindent {\bf Without mask data augmentation.}
Initially, we aligned color checkers using homography based on dataset corner locations, but imprecisions led to alignment errors at the pixel level. Our mask-based data augmentation approach eliminates reliance on specific corner positions, producing more accurate scene-harmonized color checkers that better represent the overall scene lighting.

\vspace{3pt}  \noindent {\bf Without inpainting color checker.}
In this experiment, we directly used the diffusion model to predict scene illumination RGB, not by inpainting a checker. This direct approach proves less effective than our inpainting method, highlighting the importance of color checker references for accurate illumination estimation.

\section{Conclusion}
\label{sec:conclusion}
\vspace{-1mm}


In this work, we introduce a color constancy method that leverages image-conditional diffusion models to inpaint color checkers directly into images. Our approach harnesses the rich priors of foundation models to overcome generalization challenges across varying camera sensors. By employing Laplacian decomposition, our method maintains the checker's high-frequency structure while adapting to scene illumination, enabling accurate light color estimation without camera-specific training. Experiments demonstrate robust performance in cross-camera scenarios, particularly for challenging cases, making our approach a versatile solution for real-world color constancy applications.


\newpage
\paragraph{Acknowledgements.}
This research was funded by the National Science and Technology Council, Taiwan, under Grants NSTC 112-2222-E-A49-004-MY2 and 113-2628-E-A49-023-. The authors are grateful to Google, NVIDIA, and MediaTek Inc. for their generous donations. Yu-Lun Liu acknowledges the Yushan Young Fellow Program by the MOE in Taiwan.

{\small
\bibliographystyle{ieeenat_fullname}
\bibliography{11_references}
}

\ifarxiv \clearpage \appendix \label{sec:appendix_section}

\section*{Overview}
This supplementary material presents additional details and results to complement the main manuscript. In Section \ref{sec:Implementation}, we provide comprehensive implementation details, including dataset preprocessing protocols and training configurations. Section \ref{sec:Laplacian} presents an empirical analysis of the impact of different pyramid levels in our Laplacian decomposition technique and provides implementation details of the algorithm. Section \ref{sec:Qualitative} showcases qualitative results demonstrating our method's effectiveness across various datasets and real-world scenarios. We will release our complete training and inference code along with pre-trained weights to facilitate future research in this area.

\section{Implementation Details}~\label{sec:Implementation}
\subsection{Datasets and Preprocessing}
We use two publicly available color constancy benchmark datasets in our experiments: the NUS-8
dataset~\cite{cheng2014illuminant} and the Gehler dataset~\cite{4587765}. The Gehler dataset~\cite{4587765} contains 568 original images captured by two different cameras, while the NUS-8 dataset~\cite{cheng2014illuminant} contains 1736 original images captured by eight different cameras. Each image in both datasets includes a Macbeth Color Checker (MCC) chart, which serves as a reference for the ground-truth illuminant color.


Following the evaluation protocol in \cite{afifi2019sensor}, several standard metrics are reported in terms of angular error in degrees: mean, median, tri-mean of all the errors, the mean of the lowest 25\% of errors, and the mean of the highest 25\% of errors.

\subsection{Training Details}

For all experiments, we process the raw image data before applying gamma correction for sRGB space conversion following the preprocessing protocol from \cite{hu2017fc4}. Since the pre-trained VAE was trained on sRGB images, we apply a gamma correction of $\gamma = 1/2.2$ on linear RGB images before encoding to minimize the domain gap. Conversely, after VAE decoding, we apply inverse gamma correction to convert the output back to the linear domain for metric evaluation.

All experiments are trained for 20000 iterations on an NVIDIA A6000 GPU using the Adam optimizer with an initial learning rate of $5 \times 10^{-5}$ and apply exponential learning rate decay after a 150-step warm-up period. For data augmentation, we follow FC4~\cite{hu2017fc4} to rescale images by random RGB values in [0.6, 1.4], noting that we only rescale the input images since our training does not require ground truth illumination. The rescaling is performed in the raw domain, followed by gamma correction. This is implemented through a 3×3 color transformation matrix, where diagonal elements control the intensity of individual RGB channels (color strength), and off-diagonal elements determine the degree of color mixing between channels (color offdiag). For Laplacian decomposition, we use a two-level pyramid ($L = 2$) to balance the preservation of high-frequency structural details and the suppression of low-frequency color information. Additionally, we apply local transformations to masked regions only, including brightness adjustment ($[0.8, 2.0]$), saturation adjustment ($[0.8, 1.4]$), and contrast adjustment ($[0.8, 1.4]$).

\paragraph{Three-fold Cross-validation}
For cross-validation experiments on both the NUS-8 dataset~\cite{cheng2014illuminant} and the Gehler dataset~\cite{4587765}, we use a batch size of 8. During training, we apply random crop with a probability of $p_{crop} = 0.7$, where the crop size ranges from 70\% to 100\% of the original dimensions. Color augmentation is applied with a probability of $p_{color} = 0.3$.

\paragraph{Leave-one-out Evaluation}
For the leave-one-out experiments on the NUS-8 dataset~\cite{cheng2014illuminant}, we use a batch size of 8 with gradient accumulation over 2 steps (effective batch size of 16). We apply random crop with a probability of $p_{crop} = 0.75$, where the crop size ranges from 70\% to 100\% of the original image dimensions, and color augmentation with a probability of $p_{color} = 0.65$. 

For the Gehler dataset~\cite{4587765}, when training on Canon5D and evaluating on Canon1D, we use a batch size of 8, apply random crop with a probability of $p_{crop} = 0.75$ (crop size from 70\% to 100\%), and color augmentation with a probability of $p_{color} = 0.85$. Similarly, when training on Canon1D and evaluating on Canon5D, we maintain the same batch size of 8, with random crop probability of $p_{crop} = 0.7$ and crop size ranging from 50\% to 100\%, while keeping the color augmentation probability at $p_{color} = 0.85$.

\paragraph{Cross-dataset Evaluation}
When training on NUS-8~\cite{cheng2014illuminant} and testing on the Gehler dataset\cite{4587765}, we use a batch size of 8 with gradient accumulation over 2 steps (effective batch size of 16). We apply random crop with a probability of $p_{crop} = 0.75$, where the crop size ranges from 70\% to 100\% of the original dimensions, and color augmentation with a probability of $p_{color} = 0.6$. Conversely, when training on the Gehler dataset~\cite{4587765} and testing on NUS-8~\cite{cheng2014illuminant}, we use a batch size of 8 without gradient accumulation. We apply random crop with the same probability of $p_{crop} = 0.75$ and size range of 70\% to 100\%, while color augmentation is applied with a probability of $p_{color} = 1.0$.

\paragraph{SDXL Inpainting (SDEdit)}
For the SDXL inpainting model \cite{rombach2021highresolution} with LoRA fine-tuning experiments, we use a learning rate of $5 \times 10^{-5}$ and a LoRA rank of 4. In the cross-dataset experiment from the NUS-8 dataset~\cite{cheng2014illuminant} to the Gehler dataset~\cite{4587765}, we train for 20,000 iterations with batch size 4.

\subsection{Inference Settings}
\paragraph{Full Model} Following \citeauthor{garcia2024fine} \cite{garcia2024fine}, we employ DDIM scheduler with a fixed timestep $t=T$ and \textbf{trailing} strategy during inference for deterministic single-step generation. Our implementation is based on the stable-diffusion-2-inpainting model \cite{rombach2021highresolution}.
\paragraph{SDXL Inpainting (SDEdit)} 
For comparison, we also implement a version using SDXL inpainting model \cite{rombach2021highresolution} with LoRA \cite{hu2021lora} fine-tuning. During inference, we use the DDIM scheduler with 25 denoising steps and SDEdit with a noise strength of 0.6, a guidance scale of 7.5, and a LoRA scale of 1. The final illumination estimation is obtained by computing the median from an ensemble of 10 generated samples.
\section{Laplacian Decomposition}~\label{sec:Laplacian}
\begin{figure}[t]
    \centering
    \includegraphics[width=1\linewidth]{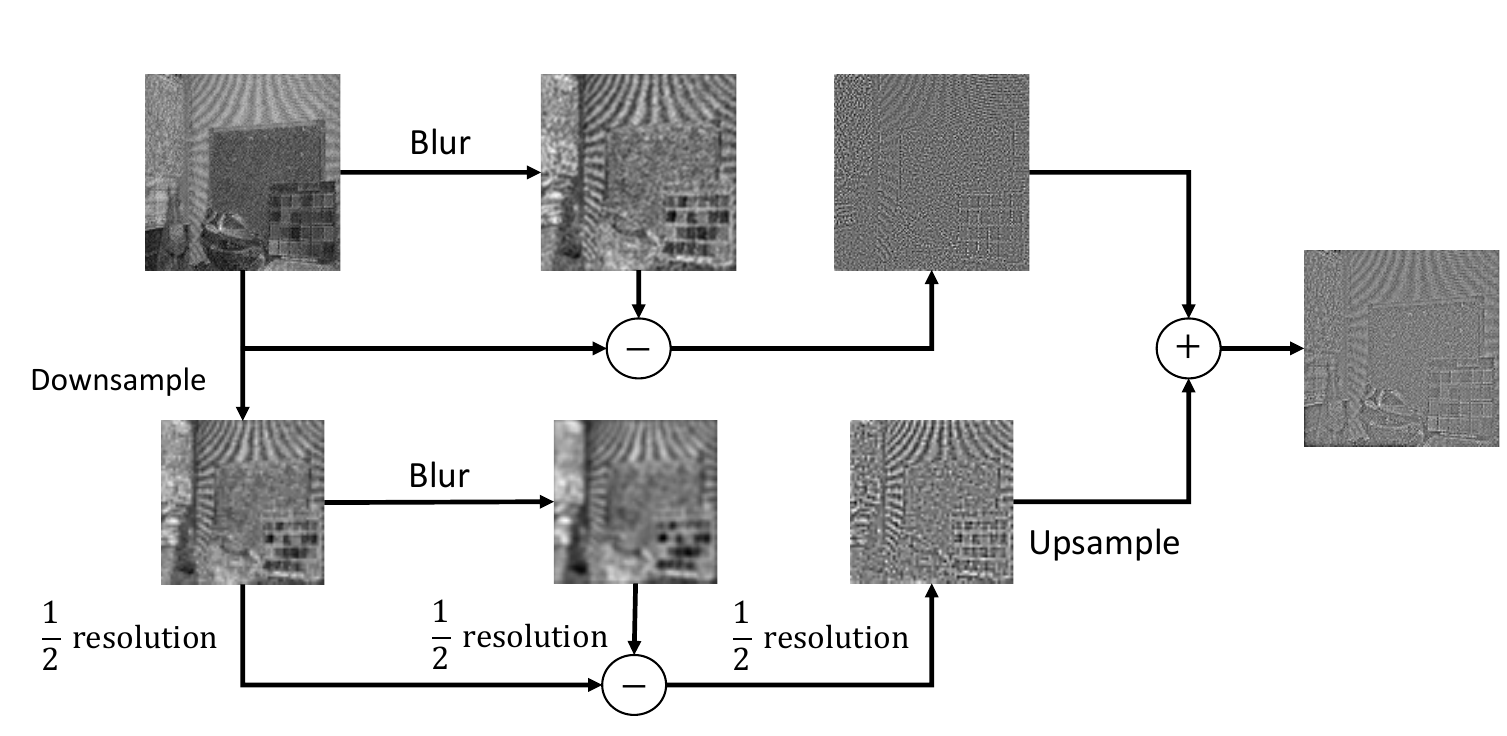}
    \caption{\textbf{Flow diagram of Laplacian decomposition.} Frequency component fusion through two-level ($1/2$ resolution) blur, downsample, and composition operations.}
    \label{fig:lapacian}
\end{figure}

\subsection{Laplacian Decomposition Visualization}
Figure~\ref{fig:lapacian} visualizes the algorithm flow of our Laplacian decomposition technique. Algorithm~\ref{alg:laplacian} outlines the detailed steps of this process, which preserves high-frequency structural details while allowing illumination-dependent color adaptation, enabling accurate scene illumination estimation.

\begin{algorithm}[h!]
\footnotesize
\SetAlgoLined
\DontPrintSemicolon
\KwIn{Input latent $z \in \mathbb{R}^{B \times C \times H \times W}$, pyramid levels $L$}
\KwOut{High-frequency components $z_h$}
\texttt{Initialize} $z_h = 0$\;
$k \leftarrow$ \texttt{3×3 Gaussian kernel}\;
\For{\texttt{each channel} $c$ \texttt{in} $C$}{
    $z_{\text{curr}} \leftarrow z[c]$ \tcp*[r]{\texttt{Current level features}}
    \For{$l = 0$ \texttt{to} $L-1$}{
        $z_{\text{blur}} \leftarrow k * z_{\text{curr}}$ \tcp*[r]{\texttt{Gaussian blur}}
        $z_{\text{high}} \leftarrow z_{\text{curr}} - z_{\text{blur}}$ \tcp*[r]{\texttt{High-freq details}}
        \eIf{$l = 0$}{
            $z_h[c] \leftarrow z_{\text{high}}$
        }{
            $z_h[c] \leftarrow z_h[c] + \texttt{Upsample}(z_{\text{high}})$
        }
        $z_{\text{curr}} \leftarrow \texttt{AvgPool}(z_{\text{blur}})$ \tcp*[r]{\texttt{Downsample}}
    }
}
\Return $z_h$
\caption{High-frequency Extraction via Laplacian Pyramid}
\label{alg:laplacian}
\end{algorithm}

\subsection{Analysis of Pyramid Level Selection}
We conduct experiments with different numbers of pyramid levels (L = 1,2,3) to analyze the effectiveness of our Laplacian decomposition. As shown in \cref{tab:pyramid_levels}, using two-level decomposition (L = 2) achieves the best performance across all metrics. Adding more levels not only increases computational complexity but also leads to performance degradation, as the additional levels introduce more low-frequency information that can adversely affect the harmonious generation of color checkers.
\section{Additional Qualitative Results}~\label{sec:Qualitative}

\subsection{Benchmark Datasets}
On the NUS-8 dataset~\cite{cheng2014illuminant} and Gehler dataset~\cite{4587765}, we utilize the original mask locations to place fixed-size neutral color checkers in our experiments. The results \cref{fig:suppl_NUS_demo} and \cref{fig:suppl_gehler_demo} demonstrate our method's ability to generate structurally coherent color checkers that naturally blend with the scene while accurately reflecting local illumination conditions, enabling effective color cast removal across diverse lighting scenarios.

\subsection{In-the-wild Images}
For in-the-wild scenes, we adopt a center-aligned placement strategy to address camera vignetting effects, which can impact color accuracy near image edges. This consistent central positioning not only mitigates lens shading issues but also demonstrates our method's flexibility in color checker placement. The results \cref{fig:suppl_inthewild_demo} validate our approach's robustness in practical photography applications, showing consistent performance in white balance correction despite the fixed central placement strategy.
\subsection{Interactive Visualization}
We provide an interactive HTML interface that visualizes results with color checkers placed at different locations within scenes. The visualization demonstrates that our method produces accurate outputs with minimal variation across different placement positions. The results show that the estimated illumination values consistently cluster near the ground truth target regardless of the checker's position, confirming our method's reliability and position-independence in illumination estimation.

\begin{figure}[t]
    \centering
    \includegraphics[width=1\linewidth]{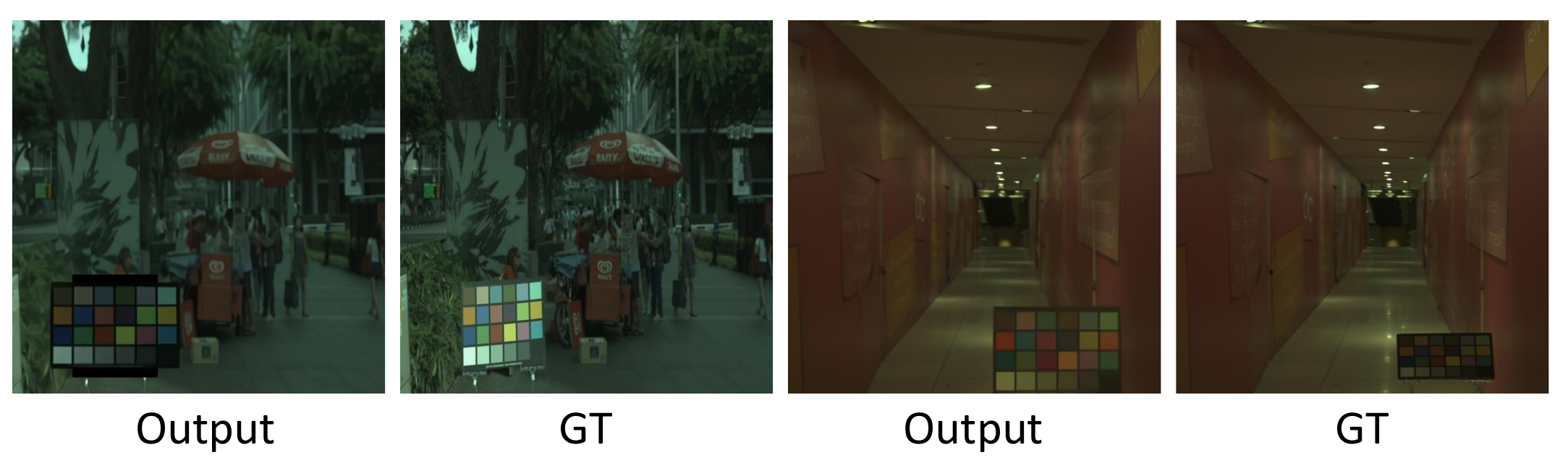}
    \vspace{-7mm}
    \caption{\textbf{Failure cases.} Our approach struggles when there is a significant mismatch between the illumination of the original color checker and the ambient lighting in the scene.}
    \label{fig:failure}
\end{figure}

\begin{table}[t]
\centering
\small
\caption{Analysis of different pyramid levels in Laplacian composition. Results are trained on the NUS-8 dataset~\cite{cheng2014illuminant} and tested on Gehler dataset~\cite{4587765} .}
\begin{tabular}{lccccc}
\toprule
Level & Mean & Median & Best-25\% & Worst-25\% &\\
\midrule
L = 1 & 3.53 & 3.27 & 1.48 & 6.03\\
L = 2 & \textbf{2.35} & \textbf{2.02} & \textbf{0.78} & \textbf{4.57}  \\
L = 3 & 3.16 & 2.83 & 1.25 & 5.62\\
\bottomrule
\end{tabular}
\label{tab:pyramid_levels}
\end{table}
\begin{figure*}[t]
    \centering
    \includegraphics[width=0.8\linewidth]{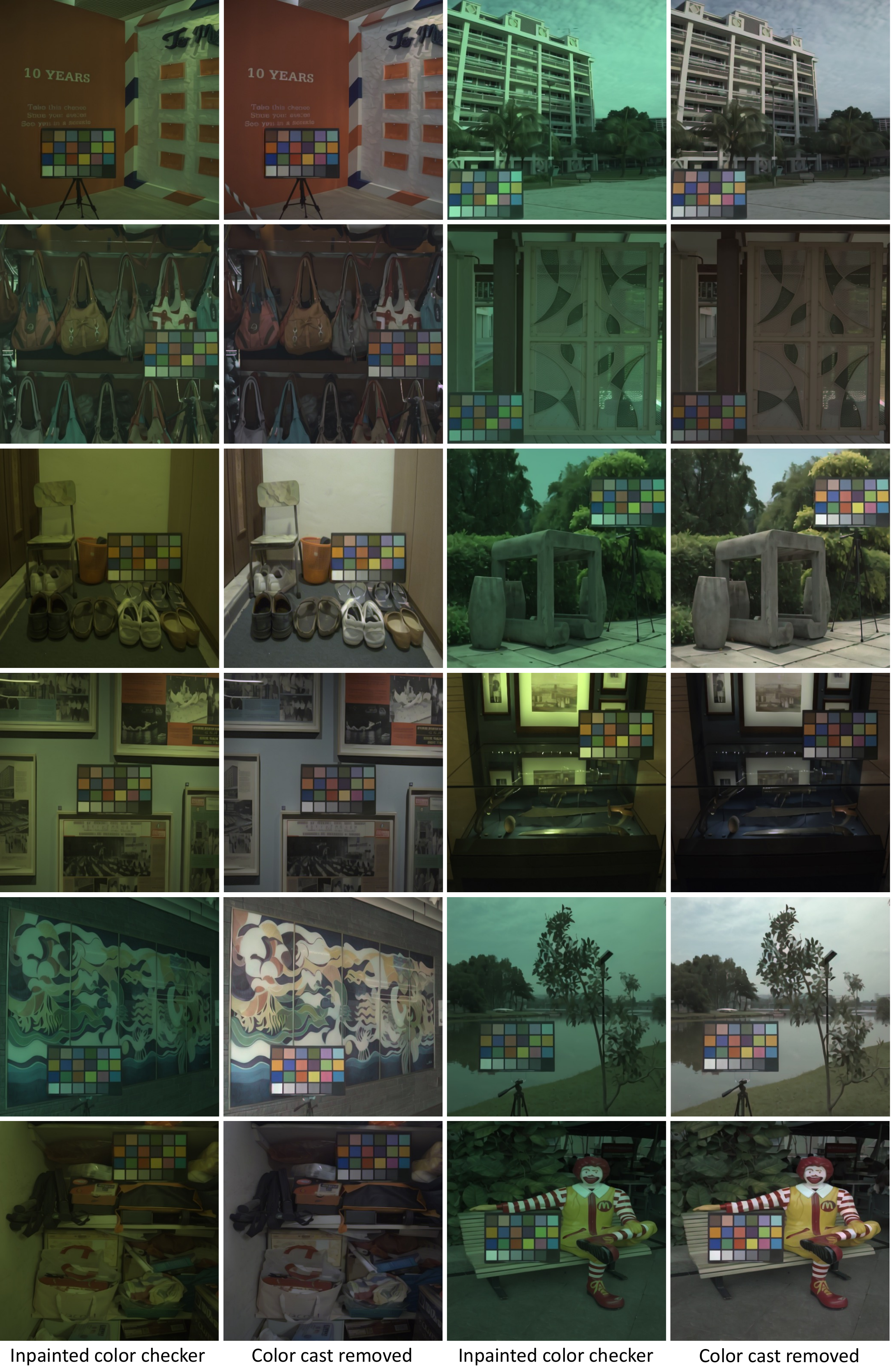}
    \caption{Qualitative results for the NUS-8 dataset~\cite{cheng2014illuminant}.}
    \label{fig:suppl_NUS_demo}
\end{figure*}
\begin{figure*}[t]
    \centering
    \includegraphics[width=0.8\linewidth]{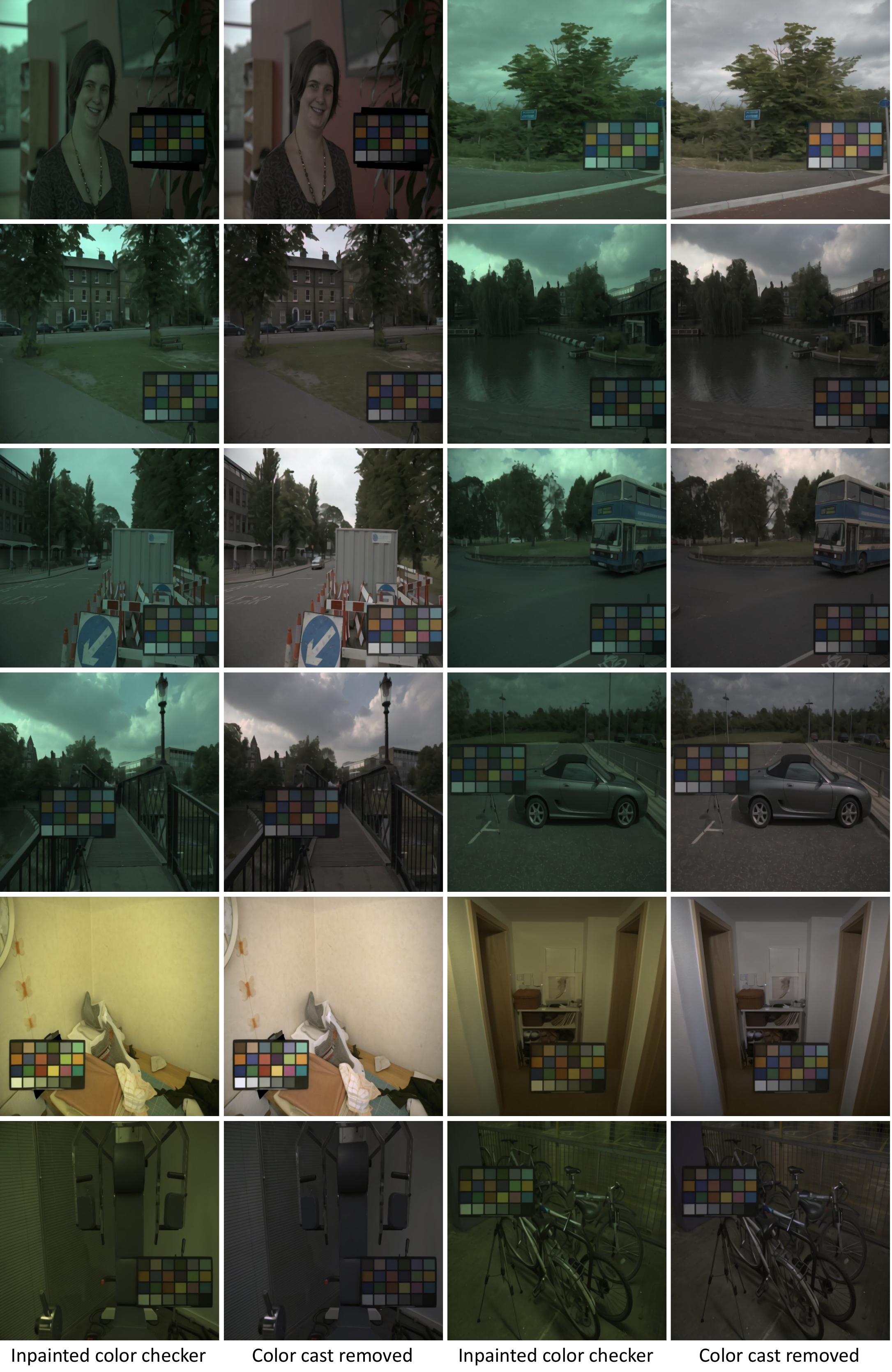}
    \caption{Qualitative results for the Gehler dataset~\cite{4587765}. }
    \label{fig:suppl_gehler_demo}
\end{figure*}
\begin{figure*}[t]
    \centering
    \includegraphics[width=0.8\linewidth]{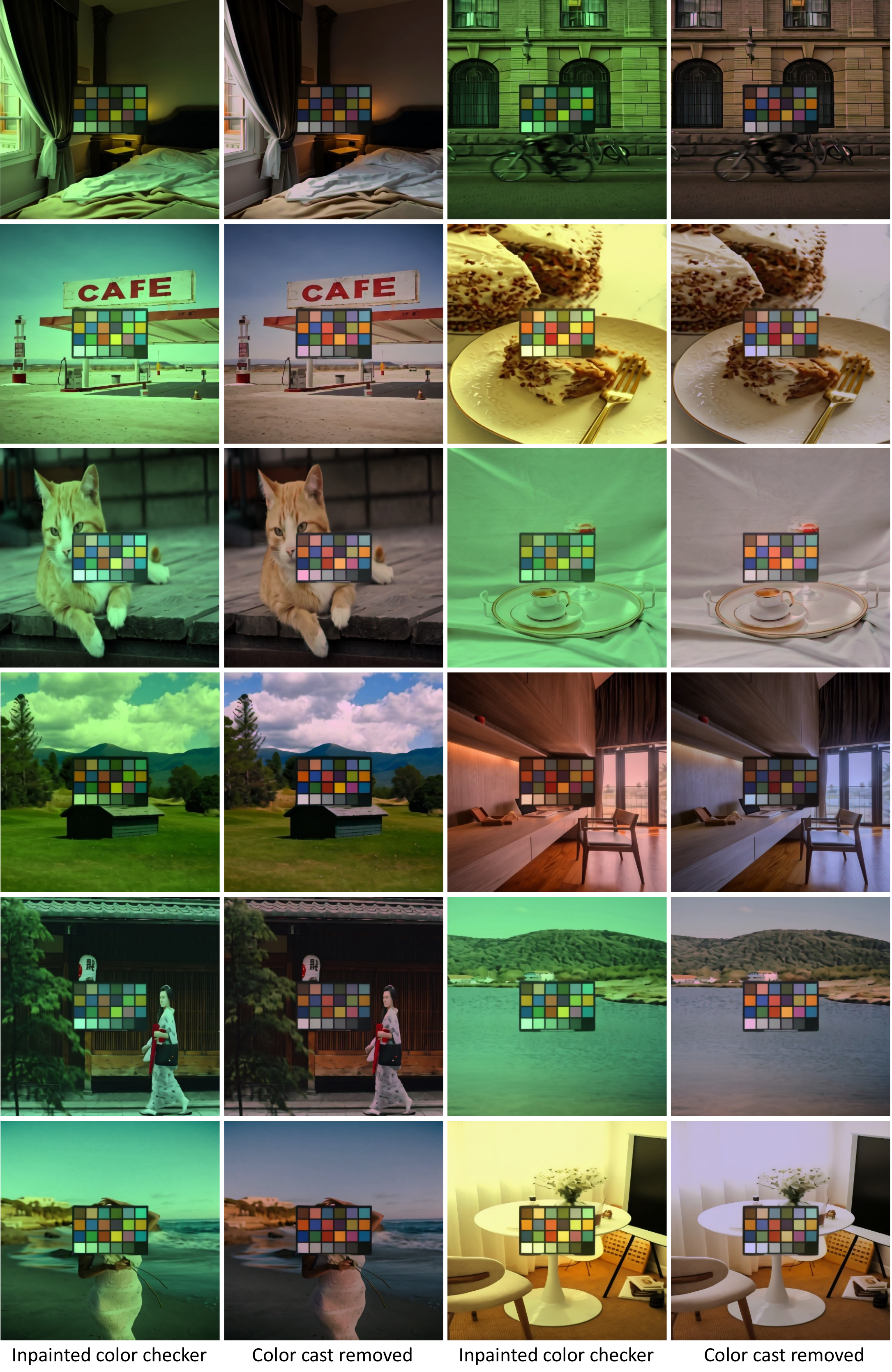}
    \caption{Qualitative results for in-the-wild images with center-placed color checkers.}
    \label{fig:suppl_inthewild_demo}
\end{figure*}

\section{Limitations}
As shown in \cref{fig:failure}, our method struggles when there is a significant mismatch between the inpainted color checker and the scene's ambient lighting. This typically occurs in challenging scenarios with multiple strong light sources of different colors or complex spatially-varying illumination. While diffusion models provide strong image priors, they sometimes prioritize visual plausibility over physical accuracy, especially in extreme lighting conditions.

Our approach also shows sensitivity to dataset size, similar to personalization effects observed in DreamBooth~\cite{ruiz2022dreambooth}. For datasets with limited samples, we need to crop smaller mask regions to ensure the model can effectively learn the color checker's appearance and structure. In our experiments, we found that when the training dataset is extremely small, the model generates color checkers with unexpected appearances and distorted structures, preventing accurate color extraction for illumination estimation. This limitation suggests potential future directions for improving our method through more efficient learning strategies or additional data augmentation techniques to better handle scenarios with limited training data. \fi

\end{document}